\documentclass[conference]{IEEEtran}
\IEEEoverridecommandlockouts
\usepackage{amsmath,amssymb,amsfonts}
\usepackage{algorithmic}
\usepackage{graphicx}
\usepackage{textcomp}
\usepackage{xcolor}
\usepackage{multirow}
\usepackage{pifont}
\usepackage{tabularx}
\usepackage[numbers,sort&compress]{natbib}
\begin{document}


\title{RefSAM3D: Adapting SAM with Cross-modal Reference for 3D Medical Image Segmentation}

\author{XiangGao, Kai Lu\thanks{Corresponding author: 961340955@qq.com}\\
{Nanjing Drum Tower Hospital, Nanjing University CN}}

\maketitle

\begin{abstract}
The Segment Anything Model (SAM), originally built on a 2D Vision Transformer (ViT), excels at capturing global patterns in 2D natural images but struggles with 3D medical imaging modalities like CT and MRI. These modalities require capturing spatial information in volumetric space for tasks such as organ segmentation and tumor quantification. To address this challenge, we introduce RefSAM3D, which adapts SAM for 3D medical imaging by incorporating a 3D image adapter and cross-modal reference prompt generation. Our approach modifies the visual encoder to handle 3D inputs and enhances the mask decoder for direct 3D mask generation. We also integrate textual prompts to improve segmentation accuracy and consistency in complex anatomical scenarios. By employing a hierarchical attention mechanism, our model effectively captures and integrates information across different scales. Extensive evaluations on multiple medical imaging datasets demonstrate the superior performance of RefSAM3D over state-of-the-art methods. Our contributions advance the application of SAM in accurately segmenting complex anatomical structures in medical imaging.

\end{abstract}


\section{Introduction} 

Medical image segmentation is a fundamental task in the field of medical imaging, primarily aimed at identifying and extracting specific anatomical structures from medical images, such as organs, lesions, and tissues. This process is crucial for numerous clinical applications, including computer-aided diagnosis, treatment planning, and monitoring of disease progression. Accurate image segmentation provides precise volumetric and shape information about target structures, which is essential for further clinical applications such as disease diagnosis, quantitative analysis, and surgical planning\cite{Obuchowicz2024Clinical,Addimulam2020Deep,Khalifa2024AI}.

Currently, recent breakthroughs in foundational models for image segmentation \cite{kirillov2023segment,Zou2023Segment} have yielded transformative results, leveraging extensive datasets to capture general representations that exhibit exceptional generalizability and performance. However, despite these strides, significant challenges arise when applying these models, particularly SAM, to medical image segmentation. For example, Huang et al. \cite{Huang2023Segment} demonstrated that SAM performs suboptimally on medical data, especially with objects that have irregular shapes or low contrast. Three main factors limit SAM's effectiveness in this domain: (1) Medical images, which often differ significantly from natural images, tend to be smaller, irregular in shape, and low in contrast, complicating direct application of the model. (2) Medical structures typically have blurred or indistinct boundaries, while SAM’s pre-training data includes predominantly well-defined edges, reducing segmentation accuracy and stability. (3) Medical imaging data often exists in three-dimensional forms with rich volumetric details, yet SAM's hint engineering was developed for two-dimensional data, limiting its ability to leverage 3D spatial features essential in medical contexts.

To enhance SAM's performance in medical imaging tasks, it is crucial to adapt and fine-tune the model to address domain-specific challenges. Recent studies have shown that parameter-efficient transfer learning (PETL) techniques, such as LoRA \cite{DBLP:journals/corr/abs-2106-09685} and Adapters\cite{DBLP:conf/emnlp/PothSPPEIVRGP23}, are effective in this context. For instance, Med-Tuning\cite{shen2024medtuningnewparameterefficienttuning} reduces the domain gap between natural images and medical volumes by incorporating Med-Adapter modules into pre-trained visual foundation models. SAMed\cite{zhang2023customizedsegmentmodelmedical} employs a low-rank approximation (LoRA) fine-tuning strategy to adjust the image encoder, prompt encoder, and mask decoder of the Segment Anything Model (SAM), achieving a balance between performance and deployment cost. However, these approaches predominantly focus on pure 2D adaptation,  not fully exploiting the three-dimensional information inherent in volumetric medical data. Nowadays, research is gradually shifting focus to better utilize the extensive data available in the 3D domain. The related methodologies can be categorized into two main approaches: one relies on prompt design based on SAM\cite{wang2024sammed3dgeneralpurposesegmentationmodels, wu2023medicalsamadapteradapting, Gong_2024}, and the other achieves fully automatic segmentation when the segmented objects exhibit relatively regular shapes and positions\cite{xie2024masksamautopromptsammask,li2024autoprosamautomatedpromptingsam}. The automatic prompt generation fails to leverage specialized medical knowledge and struggles to capture critical features due to blurred boundaries and small targets in medical images. These limitations result in suboptimal performance of automated methods, indicating further optimization.

In this paper, we propose Ref-SAM3D, an innovative approach that integrates textual prompts to enhance segmentation accuracy and consistency in complex anatomical scenarios. By incorporating text-based cues, our method enables SAM to perform referring expression segmentation within a 3D context, allowing the model to process both visual inputs and semantic descriptions for more intelligent segmentation strategies. We introduce a hierarchical attention mechanism that significantly improves the model's ability to capture and integrate information across different scales. This mechanism focuses on critical feature layers while filtering out irrelevant data, thereby enhancing segmentation precision and robustness, particularly in complex 3D structures. By integrating information across multiple scales, the model achieves a nuanced understanding of volumetric data, leading to more precise medical image segmentation. Additionally, we adapt the visual encoder to handle 3D inputs and enhance the mask decoder for direct 3D mask generation, bridging the gap between SAM's 2D architecture and the demands of 3D medical imaging. This adaptation is crucial for ensuring the model's applicability and effectiveness in this domain. We evaluate our approach on multiple medical imaging datasets, demonstrating its superior performance compared to state-of-the-art methods. Our experiments highlight the effectiveness of our model in accurately segmenting complex anatomical structures, thereby advancing the application of SAM in medical imaging. Our contributions are as follows:

1. We introduce a cross-modal reference prompt generation mechanism that integrates text and image embeddings into a unified feature space, facilitating effective cross-modal interaction.

2. We develop a hierarchical attention mechanism that significantly improves the model's ability to capture and integrate information across different scales, leading to improved segmentation precision and robustness, particularly in complex 3D structures.

3. We achieve state-of-the-art results across multiple benchmarks, demonstrating superior performance in 3D medical image segmentation tasks.

\section{Related work}
\subsection{Vision Foundation Models}
With the rapid development of foundation models in computer vision, recent research has focused on leveraging large-scale pre-training to create adaptable models with zero-shot and few-shot generalization capabilities \cite{zhang2023segmentmodelsammedical, Ma_2024,shaharabany2023autosamadaptingsammedical,na2024segmentcellsambasedautoprompting}. These Vision Foundation Models (VFMs) draw inspiration from language foundation models like GPT-series, showing remarkable adaptability across domains and tasks using pre-training and fine-tuning paradigms \cite{DBLP:journals/corr/abs-2111-01243}. Notable examples include CLIP \cite{DBLP:journals/corr/abs-2103-00020} and ALIGN \cite{DBLP:journals/corr/abs-2102-05918}, which employ image-text pairs to achieve zero-shot generalization across tasks like classification and video understanding. Building on these foundations, segmentation-specific models like SEEM \cite{DBLP:conf/nips/ZouYZLLWWGL23} and SegGPT \cite{wang2023seggptsegmentingcontext} have emerged to address more complex tasks. SEEM enhances VFM capabilities by introducing a universal prompting scheme that enables semantic-aware open-set segmentation, expanding their use in real-world scenarios. SegGPT, in turn, standardizes segmentation data and employs in-context learning for both images and videos, allowing it to handle diverse segmentation tasks without requiring additional task-specific training. Complementing these advances, DINOv2 \cite{oquab2024dinov2learningrobustvisual} scales up Vision Transformer (ViT) pre-training by increasing data and model size, producing more general and transferable visual features that simplify fine-tuning across a wide range of tasks, further broadening VFM applicability. SAM (Segment Anything Model) \cite{kirillov2023segment}  is one of the most notable VFMs for general-purpose image segmentation. Pre-trained on 11 million images and 1 billion masks, SAM enables interactive, prompt-driven zero-shot segmentation across a wide variety of visual tasks. Its impressive versatility has made it a key model for applications like image segmentation, inpainting, and tracking, though it still faces limitations in specific domains such as medical imaging, camouflage detection, and shadow segmentation \cite{wang2023detectshadowsegmentvideo}. 

\subsection{Adapting SAM in Medical Imaging}    
The adaptation of SAM for medical imaging has evolved rapidly, driven by its impressive zero-shot performance in natural image segmentation. Initial evaluation studies \cite{Deng2023Segment, He2023Accuracy, Hu2023When, Zhou2023Can} examined SAM's applicability to medical image segmentation, but its performance often fell short due to the domain gap between natural and medical images. For instance, He et al.\cite{He2023Accuracy} noted a performance gap of up to 70\% in Dice scores compared to domain-specific models. This highlighted the need for task-specific fine-tuning. Following this, research attention shifted from evaluation to the adaptation of SAM for medical images \cite{Gong_2024, li2024autoprosamautomatedpromptingsam,Ma_2024, Wu2023Medical}. Several studies have experimented with fine-tuning SAM by modifying its prompt design to handle the specific characteristics of medical data. SAM-Med2D\cite{cheng2023sammed2d}, for example, leveraged more comprehensive prompts, including points, bounding boxes, and masks, to optimize SAM for 2D medical image segmentation, while MSA\cite{Wu2023Medical} incorporated point prompts and adapters to inject medical domain knowledge into SAM’s architecture. Although these approaches enhanced SAM’s performance, the creation of prompts for each 2D slice of 3D medical data proved to be labor-intensive. Efforts to adapt SAM for 3D medical image segmentation have focused on overcoming this limitation. MedLSAM \cite{lei2024medlsamlocalizesegmentmodel} and SAM3D \cite{yang2023sam3dsegment3dscenes} applied SAM to 3D datasets, with approaches like SAMed \cite{zhang2023customizedsegmentmodelmedical} and Med-Tuning\cite{shen2024medtuningnewparameterefficienttuning} employing techniques like LoRA (Low-Rank Adaptation) to fine-tune SAM for 3D tasks. However, most of these methods have not fully addressed the critical need to account for 3D volumetric or temporal information, which is vital for medical image segmentation. Innovations such as 3DSAM-Adapter\cite{Gong_2024} and MA-SAM \cite{chen2023masammodalityagnosticsamadaptation} have incorporated 3D convolutional adapters to transform SAM’s 2D architecture into one capable of recognizing 3D structures. Similarly, SAMMed3D \cite{wang2024sammed3dgeneralpurposesegmentationmodels} introduced a framework to generate 3D prompts from 2D points, helping SAM process volumetric data more effectively. The success of these 3D adaptations highlights the importance of leveraging spatial information for more accurate segmentation. Recent trends indicate a shift towards prompt-free or semi-automatic systems, like AutoSAM Adapter \cite{li2024autoprosamautomatedpromptingsam}, which aim to maintain SAM's zero-shot capabilities while minimizing manual prompt generation.

\subsection{Parameter-Efficient Transfer Learning}
With the widespread adoption of foundational models, parameter-efficient model fine-tuning (PETL) has garnered significant attention. PETL methods can be categorized into three main groups. One approach is addition-based methods, which involve integrating lightweight adapters or prompts into the original model. These adapters or prompts allow for the fine-tuning of only a small number of additional parameters, enabling the model to adapt to specific tasks while preserving the majority of its pre-trained weights. This approach minimizes the computational overhead associated with training large models, as only the newly introduced components require optimization \cite{pan2022stadapterparameterefficientimagetovideotransfer, shen2024medtuningnewparameterefficienttuning}. Another strategy focuses on specification-based methods, which prioritize the identification and tuning of a small proportion of influential parameters from the original model. This method often employs techniques such as sensitivity analysis to determine which parameters have the most significant impact on the model's performance for a given task. By selectively updating these parameters, specification-based methods aim to achieve efficient adaptation while reducing the training burden and maintaining high performance levels \cite{zhang2023customizedsegmentmodelmedical,Gong_2024}. Additionally, reparameterization-based methods leverage low-rank representations to minimize the number of trainable parameters during the fine-tuning process. Techniques such as Low-Rank Adaptation (LoRA) and Factorized Tuning (FacT) allow models to maintain their expressive power while significantly reducing the number of parameters that need to be adjusted. This approach not only enhances efficiency but also enables strong performance across various PETL tasks, as it effectively captures the essential features required for adaptation \cite{DBLP:journals/corr/abs-2106-09685}. Recently, PETL techniques have been successfully utilized to adapt vision foundation models for a wide range of downstream tasks, including image classification, object detection, and, notably, medical image segmentation. Researchers have explored ways to fine-tune vision models efficiently while addressing the unique challenges posed by these complex tasks \cite{zhou2022conditionalpromptlearningvisionlanguage, jia2022visualprompttuning, pan2022stadapterparameterefficientimagetovideotransfer}. 

\subsection{Referring Image Segmentation}
Referring image segmentation is a task that involves segmenting a specific object in an image based on a natural language description. 
This task requires the model to understand both the visual content of the image and the semantic meaning of the text, making it a challenging problem at the intersection of computer vision and natural language processing.
With the advent of large-scale vision-language models, the performance of referring image segmentation has significantly improved.
Models like CLIP \cite{radford2021learning} and ALIGN\cite{jia2021scaling} leverage large datasets of image-text pairs to learn joint embeddings that can be used for various vision-language tasks, including referring image segmentation. These models have demonstrated strong zero-shot and few-shot capabilities, enabling them to generalize well to unseen tasks and datasets.
Recent advances have seen the adoption of transformer architectures for referring image segmentation. Transformer-based models, such as the Vision Transformer (ViT)\cite{dosovitskiy2020image}, have been adapted to this task by integrating textual information into the visual processing pipeline. Ding et al.\cite{ding2021vision} introduced a Vision-Language Transformer (VLT) approach that leverages transformer and multi-head attention mechanisms to establish deep interactions between vision and language features, significantly enhancing holistic understanding. Similarly, cross-modal attention mechanisms have become a key component in modern referring image segmentation models. These mechanisms enable the model to effectively combine visual and textual features by computing attention scores between the two modalities. Li et al.\cite{li2024refsamefficientlyadaptingsegmenting} introduced the hierarchical dense attention module to fuse hierarchical visual semantic information with sparse embeddings to obtain fine-grained dense embeddings, and an implicit tracking module to generate a tracking token and provide historical information for the mask decoder.

\section{Method}
\begin{figure*}
    \centering
    \includegraphics[width=\textwidth]{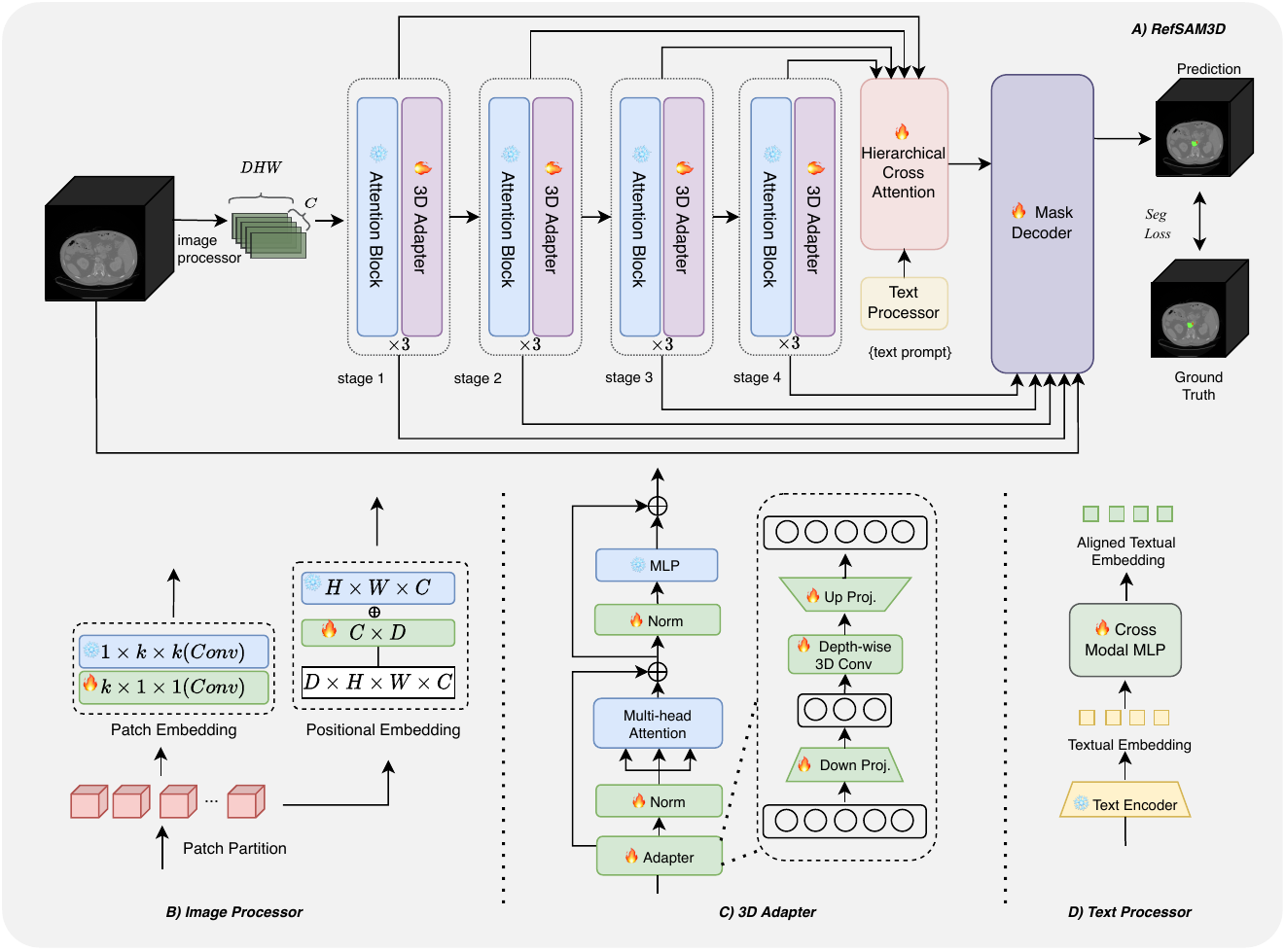}     
    \caption{(A) The overview of our proposed RefSAM3D for 3D medical image segmentation, which integrates hierarchical cross-attention between image and text modalities to generate accurate segmentation predictions. (B) The design of the Image Processor, which includes patch partitioning, convolutional-based patch embedding, and positional embedding to process volumetric 3D medical data. (C) The framework of the 3D Adapter, which incorporates multi-head attention, depth-wise 3D convolution, and up/down projection for efficient feature extraction and adaptation. (D) The pipeline of the Text Processor, which encodes textual prompts and aligns them with visual embeddings using a cross-modal MLP for enhanced segmentation guidance.}
    \label{fig:prompt-}
\end{figure*}

\subsection{Overview of Ref-SAM3D}
The original Segment Anything Model (SAM), built on a 2D Vision Transformer (ViT), is proficient in capturing global patterns within 2D natural images. However, its applicability is limited when it comes to medical imaging modalities such as CT and MRI, which involve 3D volumetric data. In these contexts, 3D information is essential for applications like organ segmentation and tumor quantification, as the characteristics of these structures must be captured from a 3D perspective. Relying solely on 2D views can result in reduced accuracy due to potential boundary blurring and non-standard scanning postures.

\subsection{3D Volumetric Input Processing}
The original Segment Anything Model (SAM) is based on a 2D Vision Transformer (ViT), excelling at capturing global patterns in natural 2D images. However, many widely used medical imaging modalities, such as CT and MRI, are inherently three-dimensional. In applications like organ segmentation and tumor quantification, 3D information is critical as these tasks require capturing representative patterns in volumetric space. Solely relying on 2D projections can result in reduced accuracy due to ambiguous boundaries and inconsistent scanning postures. SAM, when applied to 3D medical imaging, often struggles to capture spatial information, leading to suboptimal segmentation results fully.

Moreover, medical images differ significantly from natural images in both content and structure, demanding higher anatomical precision and detail. Directly applying segmentation models trained on natural images to medical domains thus yields limited effectiveness. To enhance SAM’s performance in medical imaging tasks, the model needs to be adapted and fine-tuned to accommodate the domain-specific challenges. We introduce a 3D image adapter to enable SAM’s processing of volumetric data.

We first modify the visual encoder to handle 3D volumetric inputs. Given a 3D medical volume $V \in \mathbb{R}^{C \times D \times H \times W}$, where $C$, $D$, $H$, and $W$ denote the channel, depth, height, and width, respectively, we extract 3D features via the following steps:

\begin{itemize}

    \item \textbf{Patch Embedding}: We approximate a  $k \times k \times k $ convolution (with $k = 14$) by employing a combination of $1 \times k \times k$ and $k \times 1 \times 1$ 3D convolutions. The $1×k×k$ convolution is initialized with pre-trained 2D convolution weights, which remain frozen during fine-tuning. To manage the complexity of the model, we apply depthwise convolutions for the newly introduced $ k \times 1 \times 1 $convolutions, reducing the number of parameters that require tuning.

    \item \textbf{Positional Encoding}: In the pre-trained ViT model, we introduce an additional learnable lookup table with dimensions $(C \times D)$ to encode the positional information for 3D points $(d, h, w)$. By summing the positional embedding from the frozen $(h, w)$ table with the learnable $d$-axis embedding, we provide accurate positional encoding for the 3D data.

    \item \textbf{Attention Block}: The attention block is directly adjusted to accommodate 3D features. For 2D inputs, the query size is $[B, HW, C]$, which is easily modified to $[B, DHW, C]$ for 3D inputs while retaining all pre-trained weights. We adopt a sliding window mechanism, similar to that in the Swin Transformer, to mitigate memory overhead resulting from the increased dimensionality, optimizing the model's performance and memory footprint.

    \item \textbf{Bottleneck}: As in other studies, we enhance the bottleneck layer to better adapt to 3D tasks. Specifically, we replace 2D convolutions with 3D ones and train these layers from scratch to improve performance. To avoid the computational expense of fully fine-tuning a 3D ViT, we employ a lightweight adapter for efficient fine-tuning. The adapter comprises a down-projection and an up-projection linear layer, formulated as:
\begin{equation}
\text{Adapter}(X) = X + \text{Act}(X W_{\text{Down}}) W_{\text{Up}}
\end{equation}

where $X \in \mathbb{R}^{N \times C}$ represents the input feature, $W_{\text{Down}} \in \mathbb{R}^{C \times N'}$ and $W_{\text{Up}} \in \mathbb{R}^{N' \times C}$ are the down-projection and up-projection layers, and $\text{Act}(\cdot)$ is the activation function. Additionally, we incorporate depthwise convolutions after the down-projection layer to enhance 3D spatial awareness.

\end{itemize}
\subsection{Cross-Modal Reference Prompt Generation}

\subsubsection{Text Encoder}
Within the Segment Anything Model (SAM) framework, we carefully designed a text encoder to process textual prompts related to image segmentation tasks. Specifically, we employed the text encoder from the CLIP model, which can convert input textual prompts, such as "perform liver segmentation," into corresponding text embedding vectors.

The textual prompt is first tokenized into a sequence of tokens $T = {t_l}_{l=1}^L$. These tokens are then input into the CLIP text encoder to obtain the final embedding representation. The output of the text encoder can be expressed as:
\begin{equation}
    \mathcal{F}_e = \mathcal{E}_t(T) \in \mathbb{R}^{L \times C_e} 
\end{equation} 
Here, $\mathcal{F}_e$ is the sequence of $L$ word embeddings, each with $C_e$ dimensions, i.e., $\mathcal{F}_e = {f_i}_{i=1}^L$, where each word is represented by a $C_e$-dimensional embedding. By applying a pooling operation over these word embeddings, we obtain a sentence-level embedding $F_{e}^s \in \mathbb{R}^{C_e}$. 

\subsubsection{Cross-modal Projector}
\begin{figure*}
    \centering
    \includegraphics[width=\textwidth]{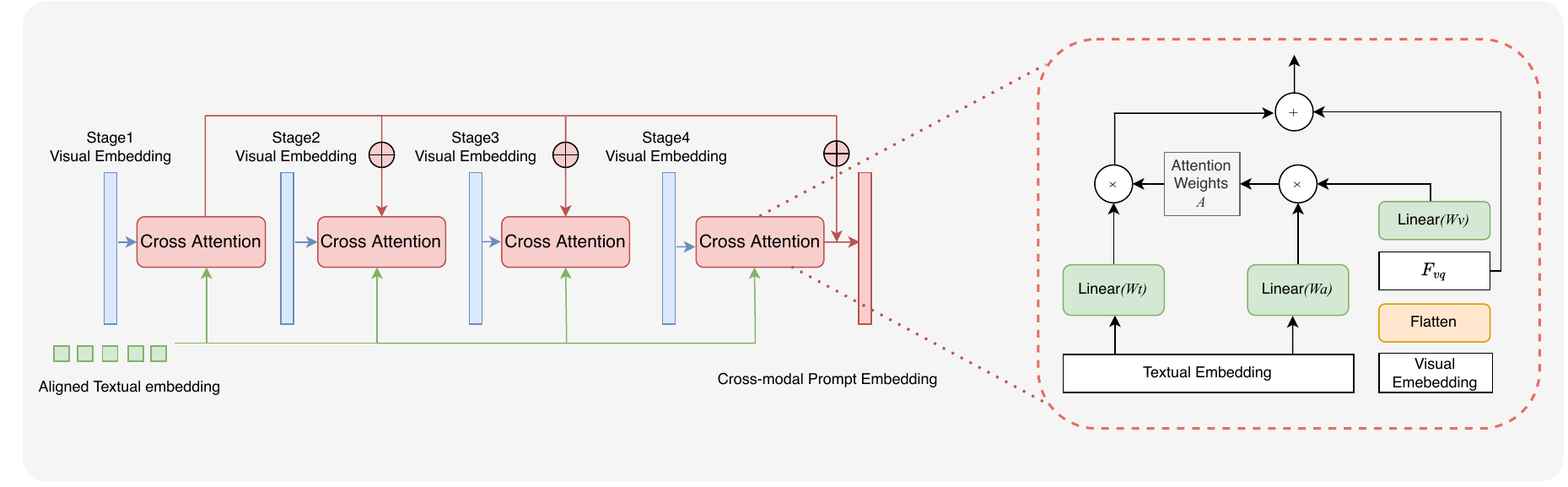}
    \caption{The structure of the Cross-Modal Prompt Embedding module. 1) The left part illustrates the overall architecture, where hierarchical visual embeddings from four stages interact with aligned textual embeddings using cross-attention mechanisms to generate cross-modal prompt embeddings. 2) The right part details the cross-attention mechanism, showing how attention weights are computed to align textual and visual embeddings through linear transformations and fusion, enabling effective multi-modal integration for downstream tasks.}
    \label{fig:prompt-小}
\end{figure*}
While text embeddings derived from pre-trained language models capture rich semantic representations, a significant gap exists between these representations and those obtained from visual encoders. This semantic disparity poses challenges in cross-modal fusion, as the two modalities do not naturally reside in the same embedding space. To address this, we adopt a strategy inspired by ViLBERT, wherein we employ an MLP to align the text and image embeddings. This allows both modalities to be projected into a unified feature space, enabling more effective interaction. Specifically, for each word embedding $f_i$ in $\mathcal{F}_e$, the sparse embedding can be obtained by adopting the cross-modal MLP:
\begin{equation}
f_{i}^s = \textit{MLP}(f_i) \in \mathbb{R}_{C_v}
\end{equation}

\subsubsection{Image feature Extraction}
As previously mentioned, we have integrated lightweight adapters into our 3D SAM to efficiently adapt the model for processing volumetric medical images. In this step, we extract the features produced by each attention block as cross-attention visual hierarchical features. 

Let $V_{i} \in \mathbb{R}^{B \times D_{i}H_{i}W_{i} \times C}$ denote the output of the $i^{th}$ attention block, where $B$ is the batch size, and $H_{i}$, $W_{i}$, and $D_{i}$ represent the height, width, and depth of the feature maps, respectively. This extraction allows us to leverage the unique focus of each attention block on different aspects of the input data, capturing a rich representation of 3D spatial patterns.

The adapted features are computed as:
\begin{equation}
V'_{i} = Adapter_{i}(V_{i}), \quad \forall i \in \{1, 2, \ldots, N\}
\end{equation}

where $N=4$. We can obtain a collection of image features:
\begin{equation}
    V' = [V'_{1}, V'_{2}, \ldots, V'_{N}].
\end{equation}

\subsubsection{Hierarchical Cross-Attention} 
~~

\textbf{Architecture: } The Hierarchical Cross-Attention architecture is designed to integrate multi-level visual features with textual inputs, enabling a deeper understanding of cross-modal data in 3D tasks such as medical image analysis. By extracting hierarchical features from each attention block in a 3D SAM, the architecture leverages the fact that each layer focuses on different aspects of the input data, from low-level details to high-level semantics. This structure enhances the model's ability to relate complex 3D spatial patterns with corresponding textual prompts, improving cross-modal understanding.

In this architecture, the inputs include both the hierarchical image features, $V' = [V'_{1}, V'_{2}, \ldots, V'_{N}]$, derived from each attention block, and a textual prompt $T$, which encodes the semantic information. These inputs are fused through a cross-attention mechanism where each layer of visual features interacts with the textual input, allowing for mutual enrichment of modalities. The output is a cross-modal prompt that combines visual and textual information, which can be fed into SAM's prompt encoder to guide tasks such as segmentation or object detection in 3D medical images.

\textbf{Cross-Attention Mechanism:} In the Hierarchical Cross-Attention architecture, the cross-attention mechanism is designed to facilitate interaction between the hierarchical image features and the textual prompt. As mentioned above, $V'_{i}$ represents the adapted feature maps extracted from the $i^{th} $ attention block and the textual prompt is $T \in \mathbb{R}^{B \times L \times C}$. 

The cross-attention process can be formally expressed as follows. For each hierarchical feature $F'_{i}$, we compute the attention scores $A_{i}$  with respect to the text $T$:
\begin{equation}
A_{i} = \text{softmax}\left(\frac{Q_{i}K^{T}}{\sqrt{d_k}}\right),
\end{equation}

where $Q_{i} \in \mathbb{R}^{B \times D{i}H_{i}W_{i}\times C}$ are the queries derived from $F'_{i}$, and $K \in \mathbb{R}^{B \times L \times C}$ are the keys derived from the textual prompt $T$. The dimensionality $d_k$ represents the size of the keys, which is a scaling factor to ensure stable gradients during training. The attention output $O_{i}$ for each feature block can then be computed as:
\begin{equation}
    O_{i} = A_{i}V_{i}
\end{equation}
where $V_{i}$ denotes the values corresponding to $F'_{i}$ and is similarly dimensioned as $F'_{i}$. The final output from the cross-attention mechanism can be represented as:
\begin{equation}
    O = [O_{1}, O_{2}, \ldots, O_{N}] \in \mathbb{R}^{B \times DHW \times C}
\end{equation}
resulting in a combined output that integrates both visual and textual information across multiple layers. This enriched representation is then utilized as a cross-modal prompt in the subsequent stages of SAM’s prompt encoder, effectively bridging the gap between visual features and semantic understanding derived from text.

\subsection{LightWeight Mask Decoder}
The original SAM mask decoder comprises merely two transformer layers, two transposed convolution layers, and a multilayer perception layer. In the context of 3D medical image processing tasks, we have replaced the 2D convolutions with 3D convolutions to enable direct 3D mask generation. Given that many anatomical structures or lesions in medical images are relatively small, it is often necessary to achieve higher resolution images to ensure better distinction of the segmented elements.

In the image encoder of the Segment Anything Model (SAM), the patch embedding process of the transformer backbone embeds each 16×16 patch into a feature vector, resulting in a 16×16 downsampling of the input. The SAM mask decoder employs two consecutive transposed convolution layers to upsample the feature map by a factor of four. However, the final prediction generated by SAM still has a resolution that is four times lower than the original input shape. To address this problem, we have employed progressive upsampling making moderate adjustments to the SAM decoder by integrating two additional transposed convolution operations. With each layer up-samples the feature maps by a factor of 2, the 4 transposed convolutional layers progressively restore feature maps to their original input resolution. Additionally, we introduced a Multi-Layer Aggregation Mechanism (MLAM), designing a network akin to a "U-shaped" architecture. We combined intermediate feature maps from stages 1-4 during the image encoder phase with prompts generated during the Cross-Modal Reference Prompt generation phase to enrich the mask features. To better leverage information from the original resolution, after upsampling the mask feature map to the original resolution, we concatenate it with the original image and use another 3D convolution to fuse the information and generate the final mask.

\section{Experiments} 

\subsection{Experimental Setup}
We have conducted a comprehensive evaluation of our segmentation method across four medical image segmentation tasks, encompassing three distinct imaging modalities: CT-based tumor segmentation, MRI-based cardiac segmentation, and multi-organ segmentation from multi-modal datasets. Our approach was rigorously compared against state-of-the-art methods on CT imaging tasks. Additionally, we assessed our method's performance on MRI cardiac segmentation and multi-organ segmentation tasks, providing a thorough analysis of its generalization capabilities and conducting an in-depth ablation study to elucidate the contributions of its constituent components.

\begin{table*}[t!]
\centering
\caption{Comparison with classical medical image segmentation methods on four tumor segmentation datasets.}
\renewcommand\arraystretch{1.5}
\centering
\setlength{\tabcolsep}{4.18mm}{
\begin{tabular}{c|cc|cc|cc|cc}
\hline
\multirow{2}{*}{Methods}       & \multicolumn{2}{c|}{Kidney Tumor} & \multicolumn{2}{c|}{Pancreas Tumor} & \multicolumn{2}{c|}{Liver Tumor} & \multicolumn{2}{c}{Colon Cancer} \\ \cline{2-9} 
& DICE $\uparrow$ & NSD $\uparrow$  & DICE $\uparrow$  & NSD $\uparrow$  & DICE$\uparrow$  & NSD\%$\uparrow$  & DICE$\uparrow$ & NSD $\uparrow$ \\ \hline
nnU-Net  & 73.07  & 77.47   & 41.65  & 62.54 & 60.10 & 75.41 & 43.91 & 52.52   \\
Swin-UNETR & 65.54 & 72.04 & 40.57  & 60.05 & 50.26  & 64.32 & 35.21 & 42.94  \\
UNETR++ & 56.49  & 60.04 & 37.25  & 53.59  & 37.13 & 51.99 & 25.36  & 30.68  \\
nnFormer & 45.14  & 42.28 & 36.53 & 53.97 & 45.54 & 60.67 & 24.28 & 32.19 \\
3D UX-Net & 57.59 & 58.55  & 34.83  & 52.56  & 45.54 & 60.67 & 28.50  & 32.73 \\ \hline
SAM-B(10pts/slice) & 40.07 & 34.96   & 30.55 & 32.91 & 8.56 & 5.97  & 39.14 & 42.70  \\
3DSAM-adapter(10pts/volume) & 74.91  & 84.35  & 57.47 & 79.62 & 56.61  & 69.52 & 49.99  & 65.67 \\
MA-SAM(1 relaxed 3D bbx/slice) & 93.38 & 98.91 & 80.30 & 97.19  & 75.23 & 92.31 & 65.45 & 81.40          \\
Ref-SAM3D  & \textbf{95.53}  & \textbf{99.45}  & \textbf{82.42}   & \textbf{98.41}   & \textbf{80.10}  & \textbf{93.23} & \textbf{70.14}  & \textbf{88.90} \\ \hline
\end{tabular}}
\label{table:single tumor}
\end{table*}

\begin{figure*}[t]

    \centering
    \includegraphics[width=\linewidth]{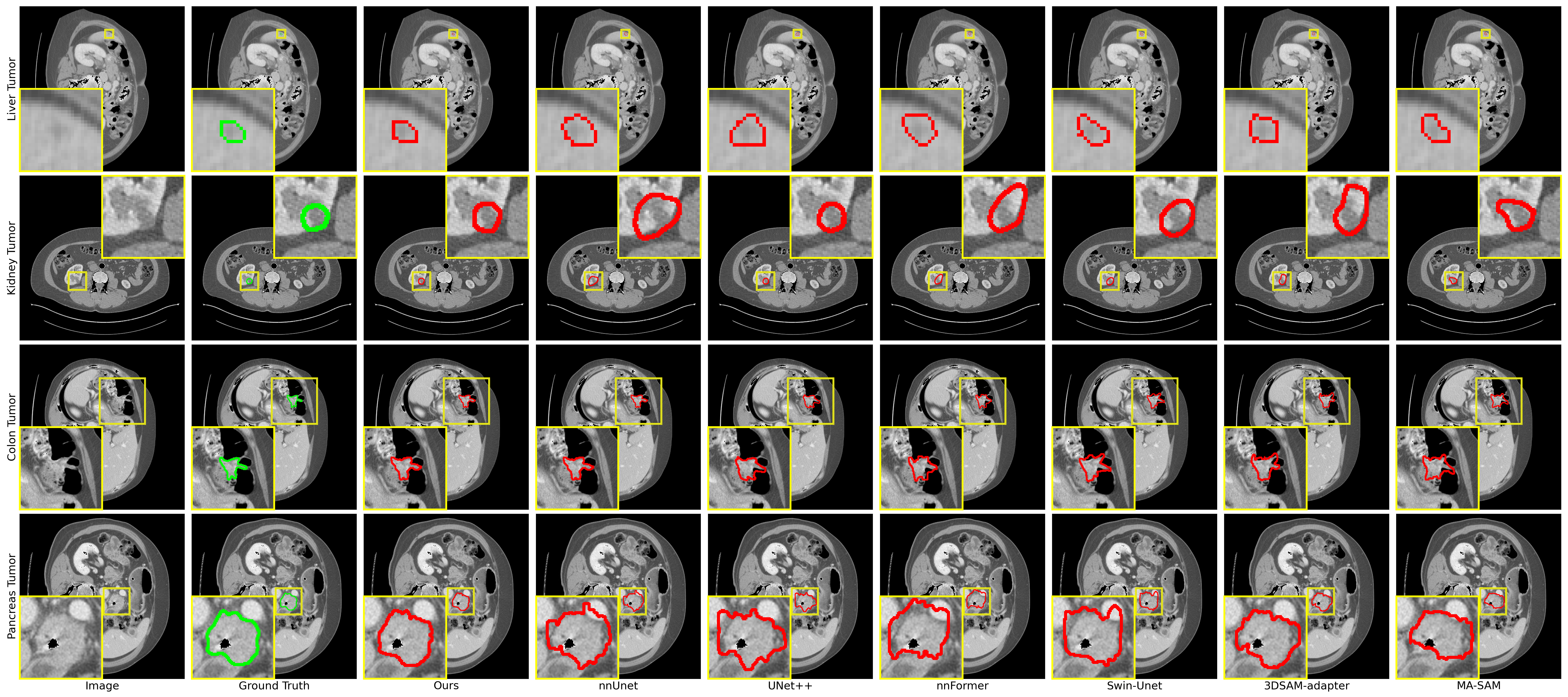}
    \centering
    \caption{Qualitative visualizations of the proposed method and baseline approaches on liver tumor, kidney tumor, pancreas tumor and colon cancer segmentation tasks.}
    \label{fig:single tumor}
\end{figure*}

\subsubsection{Datasets}
~
\begin{table}[!t]
{\small
\caption{DATASETS USED IN OUR EXPERIMENTS AND THEIR CORRESPONDING PROMPT CONTENT DESCRIPTIONS.}
\label{tab:segmentation_datasets}
\begin{tabularx}{\columnwidth}{p{1.5cm}|p{1.5cm}|X}
\hline
\textbf{Task} & \textbf{Dataset Name} & \textbf{Prompt Content} \\ 
\hline
Kidneys Tumor Segmentation & KiTS21 Challenge & CT images, kidneys, tumors, and cysts segmentation, spacing (0.5, 0.44, 0.44) to (5.0, 1.04, 1.04), dimensions (29, 512, 512) to (1059, 512, 796). \\
\hline
Pancreas Tumor Segmentation & MSD Pancreas & CT images, pancreas tumor segmentation, resolution 512×512, slices 37 to 751. \\
\hline
Liver Tumor Segmentation & LiTS Dataset & CT images, liver tumor segmentation, axial resolution 0.56-1.0 mm, z-direction resolution 0.45-6.0 mm. \\
\hline
Colon Cancer Segmentation & MSD-Colon Dataset & CT images, colon cancer segmentation, abdominal scans. \\
\hline
MRI Cardiac Segmentation & MM-WHS Challenge Dataset & MRI images, cardiac structure segmentation (LVC, RVC, LAC, RAC, AA), resolution 512×512, voxel spacing 0.3-0.6 mm. \\
\hline
Abdominal Multi-Organs Segmentation & BTCV Challenge Dataset & CT images, abdominal organ segmentation (13 organs), slice thickness 2.5-5.0 mm, in-plane resolution 0.54×0.54 mm² to 0.98×0.98 mm². \\
\hline
Multi-Modality Abdominal Multi-Organ Segmentation & AMOS 22 Dataset & CT and MRI images, abdominal organ segmentation (15 organs), varying modalities and resolutions. \\
\hline
\end{tabularx}
}
\end{table}

\textbf{Kidneys Tumor Segmentation} The KiTS21 dataset\cite{heller2023kits21challengeautomaticsegmentation} is a comprehensive collection designed for the segmentation of kidneys, tumors, and cysts in CT imaging. It comprises 300 publicly available training cases and 100 withheld testing cases. The dataset is formatted in 3D CT with files stored in the .nii.gz format. The image dimensions exhibit significant variability, with spacing ranging from (0.5, 0.44, 0.44) to (5.0, 1.04, 1.04) and sizes ranging from (29, 512, 512) to (1059, 512, 796). The dataset includes annotations for three anatomical structures: kidneys, tumors, and cysts. These structures are consistently present across all training cases, with cysts appearing in 49.33\% of the cases. This dataset serves as a critical resource for advancing automated segmentation techniques in medical imaging analysis.

\textbf{Pancreas Tumor Segmentation} The MSD Pancreas\cite{Wu2023Medical} Tumor dataset consists of 281 contrast-enhanced abdominal CT scans with annotations for both the pancreas and pancreatic tumors. This dataset is part of the Medical Segmentation Decathlon (MSD) pancreas segmentation challenge. Each CT volume has a resolution of 512 x 512 pixels, with the number of slices per scan ranging from 37 to 751. The authors filtered the dataset to retain only the axial view images containing more than 5\% pancreatic content. Consistent with previous studies, we merged the pancreas and pancreatic tumor masks into a single entity for segmentation.

\textbf{Liver Tumor Segmentation} The LiTS (Liver Tumor Segmentation Benchmark)\cite{Bilic_2023} dataset is a publicly available benchmark dataset focused on liver and liver tumor segmentation. It was created to evaluate and compare the performance of automated liver and liver tumor segmentation algorithms. The LiTS dataset comprises 201 abdominal CT scans, of which 194 contain liver lesions. The dataset is divided into 131 training cases and 70 testing cases. The resolution and quality of the CT images vary, with axial resolutions ranging from 0.56 mm to 1.0 mm and z-direction resolutions ranging from 0.45 mm to 6.0 mm.

\textbf{Colon Cancer Segmentation} The MSD-Colon Dataset\cite{Antonelli2022Medical} is a publicly available benchmark dataset focused on primary colon cancer segmentation from CT images. The dataset consists of 190 abdominal CT scans in total, which is divided into 126 training cases and 64 testing cases. Each case is annotated with segmentation masks identifying the primary colon cancer regions.

\textbf{MRI Cardiac Segmentation}
For cardiac segmentation, we utilized the Multi-Modality Whole Heart Segmentation (MM-WHS) Challenge 2017 dataset\cite{zhuang2019evaluationalgorithmsmultimodalityheart}, which contains 20 CT and 20 MRI scans with pixel-level ground-truth annotations. These scans were collected in a real clinical setting and include five anatomical labels: left ventricle blood cavity (LVC), right ventricle blood cavity (RVC), left atrium blood cavity (LAC), right atrium blood cavity (RAC), and ascending aorta (AA). In our experiments, only the CT scans were used, which contain between 177 and 363 slices, each with a resolution of 512×512 pixels and voxel spacing ranging from 0.3 to 0.6 mm.

\textbf{Abdominal Multi-Organs Segmentation}
The Beyond the Cranial Vault (BTCV) challenge dataset \cite{Landman2015MICCAI} comprises 30 CT volumes, each manually labeled with 13 different abdominal organs. The number of slices per scan ranges between 85 and 198, with a slice thickness varying between 2.5 mm and 5.0 mm. All scans have an axial resolution of 512 × 512, while the in-plane resolution varies from 0.54 × 0.54 mm² to 0.98 × 0.98 mm². We follow the data split proposed by \cite{tang2022selfsupervisedpretrainingswintransformers}, utilizing 24 cases for training and 6 cases for testing.

\textbf{Multi-Modality Abdominal Multi-Organ Segmentation}
For evaluating the model's generalization ability, we also use the Multi-Modality Abdominal Multi-Organ Segmentation Challenge (AMOS 22) dataset \cite{ji2022amoslargescaleabdominalmultiorgan}. This dataset includes abdominal CT and MRI scans from different patients, with each scan annotated for 15 organs. In line with the approach in MA-SAM, we limit our evaluation to the 12 organs common to both the AMOS 22 and BTCV datasets. For generalization testing, we utilize 300 CT scans and 60 MRI scans from the AMOS 22 training and validation sets.

\begin{figure*}[t]

    \centering
    \includegraphics[width=\linewidth]{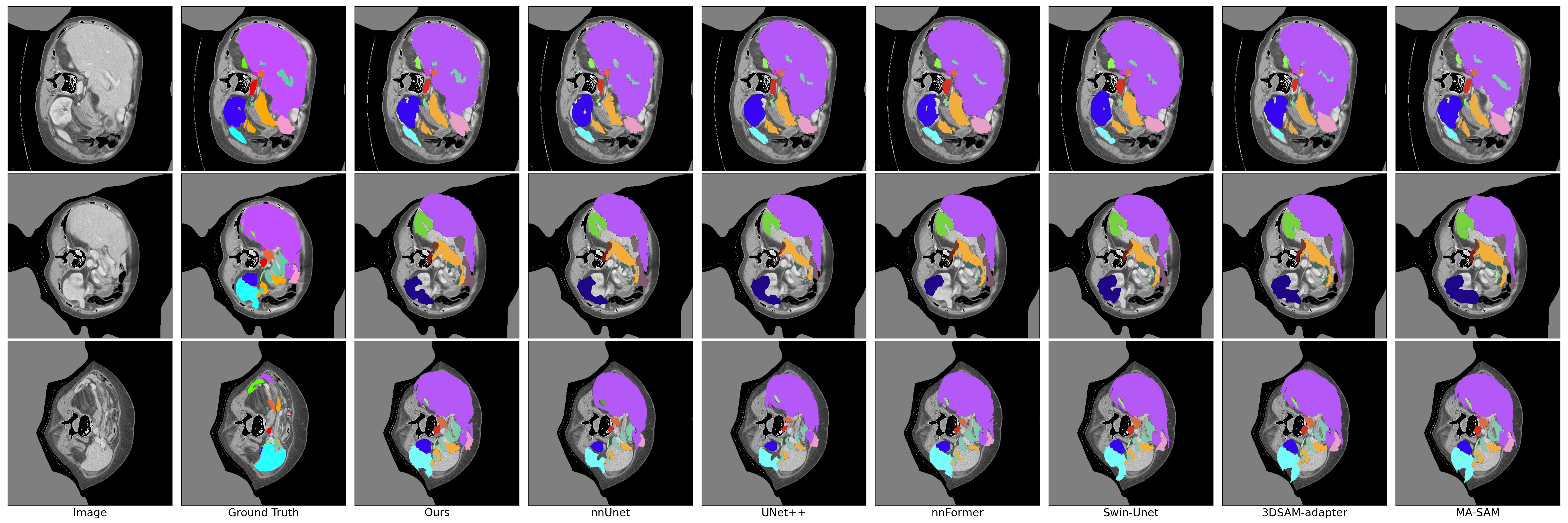}
    \centering
    \caption{Qualitative visualization of segmentation results generated from our Ref-SAM3D method and other state-of-the-art methods on BTCV dataset.}
    \label{fig:BTCV}
\end{figure*}

\begin{table*}[t]
\centering
\caption{Comparison of abdominal multi-organ segmentation results generated from our RefSAM3D method and other state-of-the-art methods on BTCV dataset.}
\renewcommand\arraystretch{1.3}
\begin{tabular}{cccccccccccccc}
\hline
\multicolumn{1}{c|}{method}     & spleen        & R.Kd          & L.Kd          & GB            & Eso.          & Liver         & Stomach       & Aorta         & IVC           & Veins         & Pancreas      & \multicolumn{1}{c|}{AG}   & Average       \\ \hline
 & \multicolumn{12}{c}{Dice{[}\%{]} $\uparrow$} & \\ \hline
\multicolumn{1}{c|}{nnU-Net}    & 97.0          & 95.3          & 95.3          & 63.5          & 77.5          & \textbf{97.4} & 89.1          & 90.1          & 88.5          & 79.0          & 87.1          & \multicolumn{1}{c|}{\textbf{75.2}} & 86.3          \\
\multicolumn{1}{c|}{Swin-UNETR} & 95.6          & 94.2          & 94.3          & 63.6          & 75.5          & 96.6          & 79.2          & 89.9          & 83.7          & 75.0          & 82.2          & \multicolumn{1}{c|}{67.3}          & 83.1          \\
\multicolumn{1}{c|}{UNETR++}    & 94.2          & 92.1          & 95.4          & 65.0            & 75.9          & 96.9          & 88.3          & 85.5          & 84.9          & 76.1          & 81.8          & \multicolumn{1}{c|}{71.3}          & 83.95         \\
\multicolumn{1}{c|}{nnFormer}   & 93.5          & 94.9          & 95.0          & 64.1          & 79.5          & 96.8          & 90.1          & 89.7          & 85.9          & 77.8          & 85.6          & \multicolumn{1}{c|}{73.9}          & 85.6          \\
\multicolumn{1}{c|}{3D UX-Net}  & 94.6          & 94.2          & 94.3          & 59.3          & 72.2          & 96.4          & 73.4          & 87.2          & 84.9          & 72.2          & 80.9          & \multicolumn{1}{c|}{67.1}          & 81.4          \\
\multicolumn{1}{c|}{3DSAM-adapter}   &  94.3 & 96.1 &  94.1 & 62.9 & 79.9 &  96.1& 83.8 & 88.4 & 85.3& 75.6& 83.1 &\multicolumn{1}{c|}{69.4} & {84.1} \\
\multicolumn{1}{c|}{MA-SAM}     & 96.7          & 95.1          & 95.4          & 68.2          & 82.1          & 96.9          & 92.8          & 91.1          & 87.5          & 79.8          & 86.6          & \multicolumn{1}{c|}{73.9}          & 87.2          \\

\multicolumn{1}{c|}{Ref-SAM3D}    & \textbf{97.1} & \textbf{94.9} & \textbf{96.1} & \textbf{70.3} & \textbf{85.2} & 97.3          & \textbf{94.1} & \textbf{92.3} & \textbf{88.8} & \textbf{80.4} & \textbf{87.5} & \multicolumn{1}{c|}{75.1}          & \textbf{88.3} \\ \hline & \multicolumn{12}{c}{HD{[}\%{]}$\downarrow$}                                                                               \\ \hline
\multicolumn{1}{c|}{nnU-Net}    & 1.07          & 1.19          & 1.19          & 7.49          & 8.56          & \textbf{1.14} & 4.84          & 14.11         & 2.87          & 5.67          & \textbf{2.31} & \multicolumn{1}{c|}{\textbf{2.23}} & 4.39          \\
\multicolumn{1}{c|}{Swin-UNETR} & 1.21          & 1.41          & 1.37          & 2.25          & 5.82          & 1.70          & 13.75         & 5.92          & 4.46          & 7.58          & 3.53          & \multicolumn{1}{c|}{3.40}           & 4.37          \\
\multicolumn{1}{c|}{UNETR++}    & 5.99          & 1.23          & 1.33          & 5.99          & 10.37         & 33.12         & 5.23          & 8.23          & 2.14          & 10.34         & 3.12          & \multicolumn{1}{c|}{2.13}          & 7.44          \\
\multicolumn{1}{c|}{nnFormer}   & 78.03         & 1.41          & 1.43          & 3.00          & 4.92          & 1.38          & 4.24          & 7.53          & 4.02          & 6.53          & 2.96          & \multicolumn{1}{c|}{2.76}          & 9.95          \\
\multicolumn{1}{c|}{3D UX-Net}  & 3.17          & 1.59          & 1.26          & 4.53          & 13.92         & 1.75          & 19.72         & 12.53         & 3.47          & 9.99          & 3.70          & \multicolumn{1}{c|}{4.11}          & 6.68          \\
\multicolumn{1}{c|}{3DSAM-adapter}  &  3.38 &   1.23 & 1.21 & 2.23 &  5.43 & 1.15 & 4.00 & 6.47 & 7.88 &  5.18 & 4.71 &  \multicolumn{1}{c|}{3.94}  &{3.90}\\
\multicolumn{1}{c|}{MA-SAM}     & \textbf{1.00} & \textbf{1.19} & 1.07          & 1.59          & 3.77          & 1.36          & 3.87          & 5.29          & 3.12          & 3.25          & 3.93          & \multicolumn{1}{c|}{2.57}          & 2.67          \\
\multicolumn{1}{c|}{Ref-SAM3D}    & 1.30          & 1.32          & \textbf{1.00} & \textbf{1.21} & \textbf{3.18} & 1.23          & \textbf{3.77} & \textbf{4.12} & \textbf{2.30} & \textbf{3.12} & 3.08          & \multicolumn{1}{c|}{2.44}          & \textbf{2.34} \\ \hline
\end{tabular}
\label{table:BTCV}
\end{table*}

\subsubsection{Implementation Details}
~

\begin{figure*}[t]

    \centering
    \includegraphics[width=\linewidth]{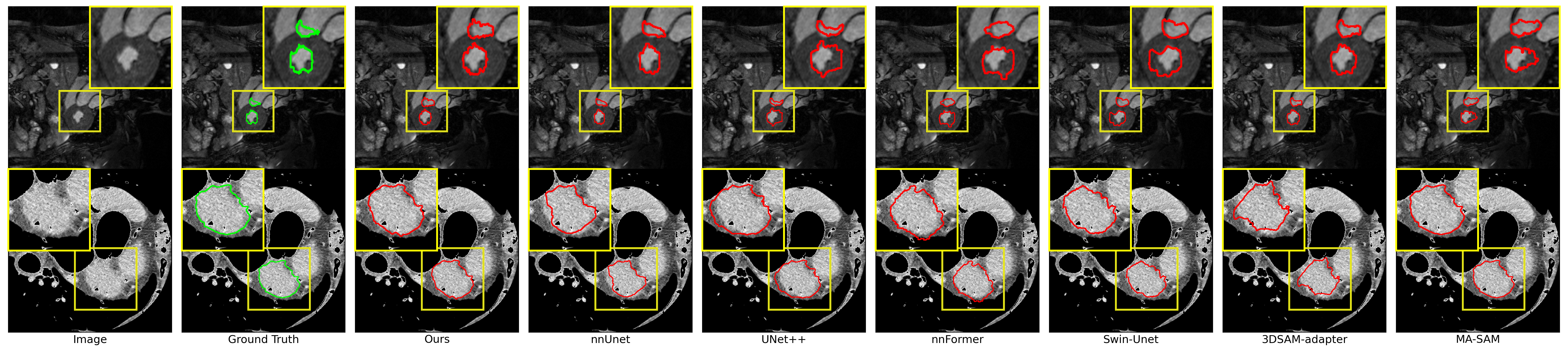}
    \centering
    \caption{Qualitative visualization of segmentation results generated from different methods for MRI cardical tumor segmentation}
    \label{fig:cardical}
\end{figure*}

We implemented our method and benchmarked it against baseline models using PyTorch and MONAI, specifically utilizing SAM\mbox{-}B for all experiments, which employs ViT\mbox{-}B as the image encoder backbone. The training was conducted on an NVIDIA A40 GPU with a batch size of $1$, using the AdamW optimizer with a linear learning rate scheduler for a total of $200$ epochs. The initial learning rate was set to $1e-4$, with a momentum of $0.9$ and a weight decay of $1e-5$. Data preprocessing involved adjusting the isotropic spacing to 1 mm. For data augmentation, we applied various transformations, including random rotation, flipping, erasing, shearing, scaling, translation, posterization, contrast adjustments, brightness modifications, and sharpness enhancements. During training, we also sampled foreground and background patches at a $1:1$ ratio. For single-organ cancer segmentation, we assessed our method's performance through comparisons with state\mbox{-}of\mbox{-}the\mbox{-}art volumetric segmentation and fine-tuning techniques, using the Dice coefficient (Dice) and Normalized Surface Dice (NSD) as evaluation metrics same as xxx \cite{}. For multi-organs segmentation, we employed Dice coefficient (Dice) and Hausdorff Distance (HD) as evaluation metrics. For each dataset, we designed specific text prompts to guide the segmentation process, as shown in Table \ref{prompt}. These prompts were carefully crafted to provide clear anatomical context while maintaining consistency across different organs and pathologies.

\subsection{Comparison with State-of-the-Arts}

Our method has been extensively evaluated against a wide range of state-of-the-art 3D medical image segmentation techniques on both CT and MRI datasets. These techniques include the CNN-based nnU-Net \cite{isensee2018nnunetselfadaptingframeworkunetbased}, an automated configuration framework evolved from the U-Net architecture \cite{DBLP:conf/miccai/RonnebergerFB15} , and the Transformer-based Swin-UNETR \cite{hatamizadeh2022swinunetrswintransformers}, which employs a hierarchical encoder structure for 3D segmentation tasks. Furthermore, we considered nnFormer \cite{DBLP:journals/tip/ZhouGZHYWY23}, a model that integrates both local and global volumetric self-attention mechanisms, and UNETR++\cite{shaker2024unetrdelvingefficientaccurate}, which enhances segmentation accuracy and efficiency through the introduction of an efficient pairing attention module. Additionally, we compared our approach with 3D UX-Net\cite{lee20233duxnetlargekernel}, a method designed to create a simple, efficient, and lightweight network that combines the capabilities of hierarchical transformers with the advantages of ConvNet modules. We also evaluated SAM-B\cite{}, which is the base model of SAM trained on natural images and directly applied to medical images without adaptation. Finally, our method was benchmarked against the latest SAM adaptation techniques, including 3DSAM-adapter\cite{Gong_2024} a promptable 3D medical image segmentation model, and MA-SAM\cite{chen2023masammodalityagnosticsamadaptation}, a framework that utilizes parameter-efficient fine-tuning strategies and 3D adapters.

The results presented in Table \ref{table:single tumor} demonstrate that our proposed Ref-SAM3D method consistently outperforms other approaches across a wide range of tasks, achieving the highest scores in nearly all scenarios, particularly excelling in challenging tumor types. In the task of kidney tumor segmentation, despite challenges such as low contrast with surrounding tissues, blurred boundaries, and high morphological heterogeneity, Ref-SAM3D achieved a DICE score of 95.53\% and an NSD of 99.45\%, surpassing other methods. For pancreatic tumors, which constitute less than 0.5\% of CT images and exhibit diverse shapes, Ref-SAM3D achieved a DICE score of 82.42\%, representing a 2.12\% improvement over existing SOTA techniques. In liver tumor segmentation, Ref-SAM3D attained a DICE score of 80.10\%, effectively handling variations in grayscale and irregular shapes. Despite the extensive distribution and complex anatomical structure of colorectal cancer lesions, Ref-SAM3D achieved a DICE score of 70.14\%, marking a 10.11\% increase over current technologies. It is noteworthy that traditional methods like nnU-Net perform well on certain tasks, yet overall, they fall short compared to newer methods such as Ref-SAM3D. Particularly when dealing with tumors that have blurred boundaries and diverse morphologies, Ref-SAM3D demonstrates significant advantages. These findings underscore the exceptional performance of Ref-SAM3D in addressing a variety of complex medical image segmentation challenge. Figure \ref{fig:single tumor} shows qualitative visualizations of theses tasks.

In the domain of multi-organ segmentation, we do experiments on BTCV dataset, the Ref-SAM3D approach demonstrated exceptional capability. Specifically,in the BTCV dataset shown in Table \ref{table:BTCV}, it achieved a DICE score of 97.1\% for the spleen, outperforming all other methods. The left and right kidneys attained DICE scores of 96.1\% and 94.9\%, respectively. The esophagus achieved a DICE score of 85.2\%, surpassing other methods, while the liver and stomach achieved scores of 97.3\% and 94.1\%, respectively. Furthermore, Ref-SAM3D showed its strengths in handling complex anatomical structures, such as the pancreas and aorta, achieving DICE scores of 87.5\% and 92.3\%, respectively. In terms of Hausdorff Distance (HD) evaluation, Ref-SAM3D also excelled, with an average HD value of 2.34, underscoring its superior boundary precision. Figure \ref{fig:BTCV} shows qualitative visualizations on BTCV tasks. From the qualitative visualization results, Ref-SAM3D demonstrates superior performance in multi-organ segmentation tasks. The method accurately identifies and segments boundaries between different organs, maintaining high segmentation precision even in cases with blurred organ boundaries or complex anatomical structures. Notably, Ref-SAM3D maintains stable segmentation performance for both small organs like the pancreas and elongated structures such as the aorta, further validating the reliability of the quantitative evaluation metrics.

In addition, in the context of cardiac tumor segmentation using MRI, as shown in Figure \ref{fig:cardical}, a qualitative assessment of predicted masks from various segmentation models indicates that our AutoSAM Adapter produces visually superior results, especially in terms of boundary precision, when compared to existing SOTA methods. 

\subsection{Generalization Evaluation}
\begin{figure}
    \centering
    \includegraphics[width=1\linewidth]{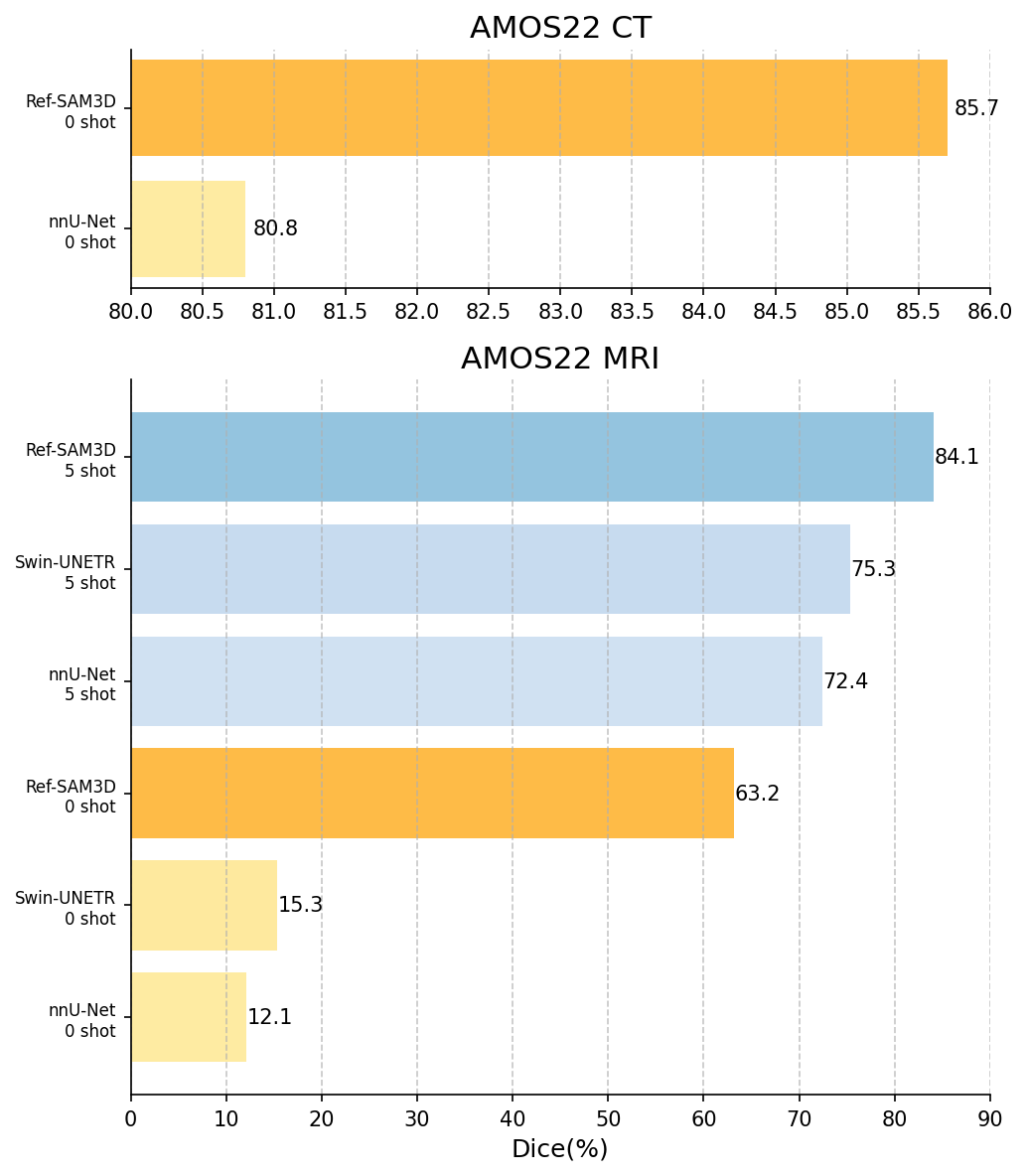}
    \caption{Comparison of zero-shot and five-shot generalization performance of Ref-SAM3D, nnU-Net and Swin-UNETR on AMOS22 CT and MRI data.}
    \label{fig:enter-label}
\end{figure}

To assess the generalization capabilities of Ref-SAM3D, we conducted comprehensive experiments across heterogeneous datasets and imaging modalities. Our evaluation framework encompassed two distinct scenarios: cross-modality generalization on the AMOS 22 dataset (comprising both CT and MRI modalities) and cross-dataset adaptation using the MM-WHS cardiac imaging dataset.

In the zero-shot generalization experiments, we evaluated the model's transferability by applying our Ref-SAM3D, trained exclusively on the BTCV CT dataset, to the AMOS 22 dataset without any additional fine-tuning. The quantitative results demonstrated remarkable performance, achieving a mean Dice coefficient of 85.7\% on CT images, indicating robust generalization across different CT acquisition protocols and patient cohorts. Notably, in the challenging cross-modality scenario of MRI segmentation, our model maintained substantial performance with a Dice score of 63.2\% ($\pm$3.1\%), significantly surpassing baseline methods including nnU-Net (12.1\%) and Swin-UNETR (15.3\%).

Furthermore, when employing a 5-shot fine-tuning strategy on the AMOS22 MRI data, Ref-SAM3D exhibited even more impressive results, achieving a Dice score of 84.1\%. This represents a substantial improvement over the fine-tuned versions of nnU-Net (72.4\%) and Swin-UNETR (75.3\%), demonstrating the model's superior adaptability and learning efficiency with minimal additional training data. These results underscore Ref-SAM3D's robust generalization capabilities and its potential as a versatile solution for medical image segmentation across different imaging modalities.

These experimental findings clearly demonstrate Ref-SAM3D's robust performance across different datasets and imaging modalities. The model's strong zero-shot generalization capabilities and impressive few-shot learning results suggest its practical value in real-world medical applications, where adapting to diverse imaging conditions with minimal additional training is essential. These characteristics position Ref-SAM3D as a promising solution for clinical deployment, particularly in scenarios requiring flexible and efficient medical image analysis tools.

\subsection{Ablation Study}
\begin{table}[t]
\renewcommand\arraystretch{1.5}
\centering
\caption{Ablation on each key component in our method.}
\label{ablation1}
\begin{tabular}{c|c|c}
\hline
                          & DICE[\%]$\uparrow$ & HD[\%]$\downarrow$   \\ \hline
Ref-SAM3D                   & 88.3 & 2.34 \\
w/o Text Prompt           & 72.3 & 7.31 \\
w/o Cross-Modal Projector & 80.1 & 4.22 \\
w/o Hierarchical Fusion   & 74.1 & 6.33 \\ \hline
\end{tabular}
\end{table}
\subsubsection{Effects of Text Prompt}
The Text Prompt in our Ref-SAM3D model provides essential semantic guidance by bridging textual descriptions and visual features, enabling better interpretation of anatomical structures. The results, as shown in Table \ref{ablation1}, without this component,  the model's performance drops significantly, with DICE score decreasing from 88.3\% to 72.3\% (-16.0\%) and HD increasing from 2.34\% to 7.31\% (+4.97\%). This substantial degradation demonstrates that the Text Prompt is crucial for leveraging linguistic context to achieve precise medical image segmentation.

\subsubsection{Effects of Cross-Modal Projector}
The Cross-Modal Projector in Ref-SAM3D plays a vital role in aligning textual and visual inputs, facilitating effective integration of multimodal information for improved segmentation. By harmonizing these inputs, the projector enhances the model's ability to utilize semantic context from text alongside visual data.   As shown in Table \ref{ablation1}, Removing this component results in an 8.2\% decrease in DICE score (from 88.3\% to 80.1\%) and HD increase from 2.34\% to 4.22\%. These results confirm When the Cross-Modal Projector is removed, the model relies on unaligned embeddings, which can lead to less effective feature integration.

\begin{table}[]
\centering
\caption{The ablation experiments of each Stage under the Hierarchical Cross-Attention}
\label{ablation2}
\renewcommand\arraystretch{1.5}
\begin{tabular}{cccc|c|c}
\hline
\multicolumn{4}{c|}{Stage}  & \multirow{2}{*}{DICE[\%]$\uparrow$} & \multicolumn{1}{c}{\multirow{2}{*}{HD[\%]$\downarrow$}} \\ \cline{1-4}
1 &2  & 3 & 4 &  &  \\ \hline
\ding{52}&\ding{52} & \ding{52} & \ding{52}   & 88.3 & 2.34  \\
\ding{52} & &  &  \ding{52} & 78.5  & 2.76  \\
 & \ding{52}  &  & \ding{52}  & 82.1  & 2.62  \\
 & & \ding{52}  & \ding{52}   & 85.4 & 2.48       \\
 &  &  & \ding{52}  & 73.78  & 2.89   \\ \hline
\end{tabular}
\end{table}
\subsubsection{Effects of Hierarchical Cross-Attention Mechanism}
The Hierarchical Fusion Mechanism in Ref-SAM3D is pivotal for integrating information across various encoder layers, enabling the model to capture detailed, multi-level semantic features essential for precise segmentation. Ablation studies, summarized in Table \ref{ablation1}, demonstrate the significance of this mechanism. Removing the Hierarchical Fusion leads to a sharp decline in segmentation accuracy, with the Dice coefficient dropping from 88.3\% to 74.1\%, and the HD increasing from 2.34\% to 6.33\%. This underscores the mechanism's role in effectively combining features across layers for better performance.

Moreover, Table \ref{ablation2} provides a systematic evaluation of each block level's contribution to the model. The results reveal that utilizing all layers (Stage 1 to 4) achieves the best performance, with a Dice score of 88.3\% and an HD of 2.34\%. In contrast, excluding specific layers leads to varied performance drops, with the shallow layers contributing significantly to contextual information and deeper layers enhancing fine-grained details. For example, when only deeper layers (Stages 3 and 4) are used, the Dice score drops to 78.5\%, and the HD increases to 2.76\%, while including only the shallow layers (Stages 1 and 2) yields a Dice score of 73.78\% and an HD of 2.89\%.

These findings underscore the necessity of a comprehensive fusion approach. Each layer's unique contributions—from the broad contextual cues in shallow layers to the detailed semantic information in deeper layers—work synergistically to enhance the model's ability to capture complex anatomical structures, ultimately improving overall segmentation accuracy and robustness.

\section{Conclusion}

We propose an effective adaptation of the Segment Anything Model (SAM) tailored for 3D medical imaging tasks, demonstrating its versatility across diverse modalities such as CT and MRI. Our framework leverages a parameter-efficient fine-tuning strategy and successfully incorporates volumetric spatial information critical for precise anatomical segmentation. By integrating hierarchical attention mechanisms and cross-modal prompt generation, our model achieves superior performance on complex segmentation tasks, significantly surpassing state-of-the-art 3D medical segmentation approaches. The proposed method exhibits remarkable generalization capability, which is essential for deploying intelligent systems across heterogeneous medical datasets. Additionally, the integration of cross-modal prompts enhances segmentation accuracy in challenging scenarios, such as tumor segmentation, where fine-grained contextual understanding is required. While our approach is highly effective, future work will focus on improving computational efficiency to enable real-time clinical applications, exploring semi-supervised learning techniques to address the challenge of limited labeled data. Overall, our method holds significant promise as a generalizable and robust segmentation framework, offering both fully automatic and promptable segmentation capabilities for a wide range of 3D medical imaging applications.


\small
\bibliographystyle{IEEEbib}
\bibliography{refs}

\vspace{12pt}
\end{document}